\documentclass{preprint}

\usepackage{xcolor}
\usepackage{booktabs}
\usepackage{graphicx}
\usepackage[colorlinks=true]{hyperref}
\usepackage{tabularx}

\usepackage[numbers,comma,sort&compress]{natbib}

\title{Generative Large Language Models Trained for Detecting Errors in Radiology Reports}
\author[1,$\dagger$]{Cong Sun}
\author[2,$\dagger$]{Kurt Teichman}
\author[1]{Yiliang Zhou}
\author[2]{Brian Critelli}
\author[2]{David Nauheim}
\author[2]{Graham Keir}
\author[2]{Xindi Wang}
\author[1]{Judy Zhong}
\author[3]{Adam E Flanders}
\author[2,*]{George Shih}
\author[1,*]{Yifan Peng}
\affil[1]{Department of Population Health Science, Weill Cornell Medicine, New York, NY}
\affil[2]{Department of Radiology, Weill Cornell Medicine, New York, NY}
\affil[3]{Department of Radiology, Thomas Jefferson University, Philadelphia, PA}
\affil[*]{Corresponding author(s). Email(s): \url{ges9006@med.cornell.edu}, \url{yip4002@med.cornell.edu}}
\affil[$\dagger$]{These authors contributed equally to this work.}

\begin{document}

\maketitle

\begin{abstract}
\textbf{Background}:
Large language models (LLMs) offer promising solutions, yet their
application in medical proofreading, particularly in detecting errors
within radiology reports, remains underexplored.

\textbf{Purpose}:
To develop and evaluate generative LLMs for detecting errors in
radiology reports during medical proofreading.

\textbf{Materials and Methods}:
In this retrospective study, a dataset was constructed with two
parts. The first part included 1,656 synthetic chest radiology reports
generated by GPT-4 using specified prompts, with 828 being error-free
synthetic reports and 828 containing errors. The second part included
614 reports: 307 error-free reports between 2011 and 2016 from the
MIMIC-CXR database and 307 corresponding synthetic reports with errors
generated by GPT-4 on the basis of these MIMIC-CXR reports and specified
prompts. All errors were categorized into four types: negation,
left/right, interval change, and transcription errors. Then, several
models, including Llama-3, GPT-4, and BiomedBERT, were refined using
zero-shot prompting, few-shot prompting, or fine-tuning strategies.
Finally, the performance of these models was evaluated using the F1 score,
95\% confidence interval (CI) and paired-sample t-tests on our
constructed dataset, with the prediction results further assessed by
radiologists.

\textbf{Results}:
Using zero-shot prompting, the fine-tuned Llama-3-70B-Instruct model
achieved the best performance with the following F1 scores: 0.769 (95\%
CI, 0.757-0.771) for negation errors, 0.772 (95\% CI,
0.762-0.780) for left/right errors, 0.750 (95\% CI,
0.736-0.763) for interval change errors, 0.828 (95\% CI,
0.822-0.832) for transcription errors, and 0.780 overall. In the
real-world evaluation phase, two radiologists reviewed 200 randomly
selected reports output by the model (50 for each error type). Of these,
99 were confirmed to contain errors detected by the models by both
radiologists, and 163 were confirmed to contain model-detected errors by
at least one radiologist.

\textbf{Conclusion}:
Generative LLMs, fine-tuned on synthetic and MIMIC-CXR radiology
reports, greatly enhanced error detection in radiology reports.
\end{abstract}


\section{Introduction}

Chest radiography is essential for diagnosing thoracic conditions~\cite{grenier1991chronic-h, bick1999pacs-f}. However, the accuracy of radiology reports can be compromised by several factors. For example, while the speech recognition software commonly used in radiology dictation efficiently transcribe spoken words, they are prone to errors, particularly when dealing with complex radiology terminology or background noise \cite{pezzullo2008voice-i, chang2011non-clinical-d, hawkins2014creation-s, minn2015improving-x}. Additionally, further errors can be introduced into radiology reports by variability in perceptual and interpretive processes, as well as cognitive biases \cite{bruno2015understanding-f, waite2017interpretive-p}.

These challenges collectively contribute to frequent errors in radiology reports, ranging from minor discrepancies to clinically significant mistakes \cite{waite2017systemic-u}. The need for accurate radiology reports cannot be overstated, as even minor errors in these reports can result in severe consequences, such as incorrect diagnoses or delayed treatments \cite{allen2017improving-r}.

Efforts to reduce radiology report errors have traditionally focused on generating more structured reports \cite{schwartz2011improving-u, larson2013improving-i, ganeshan2018structured-a, nobel2022structured-p}. However, their adoption among radiologists varies, mainly because of the rigidity of structured reports. In addition to implementing structured reports, the use of deep learning methods \cite{zech2019detecting-k, enarvi2020generating-u, chaudhari2022application-a, min2022rred-o, cai2023chestxraybert-o, nishigaki2023bert-based-w} has proven to be highly effective for detecting errors in radiology reports. While promising, these deep learning methods require substantial amounts of annotated data for training and typically focus on detecting errors at the sentence level.

Recent advancements in large language models (LLMs) provide new opportunities to address these limitations \cite{hoffmann2022training-i, wei2022chain-of-thought-q, akinciDAntonoli2024large-a}. While LLMs have shown potential in radiology applications \cite{adams2023leveraging-t, rao2023evaluating-m, gertz2023gpt-4-r}, studies evaluating their effectiveness in detecting errors in radiology reports are lacking. Additionally, proprietary LLMs (e.g., ChatGPT) have significant accessibility and implementation challenges, including high costs, data privacy concerns, and regulatory compliance issues \cite{akinciDAntonoli2024large-a}.

Although a few studies have explored the use of existing LLMs, such as GPT-4, for detecting errors in radiology reports \cite{gertz2024potential-p, schmidt2024generative-g}, research on evaluating the performance of radiology-specific fine-tuned LLMs for proofreading errors remains in its early stages. For example, Gertz et al. focused on evaluating the potential of GPT-4 for error detection in radiology reports but did not compare its performance with that of other LLMs, particularly those that are locally fine-tuned \cite{gertz2024potential-p}. Schmidt et al. evaluated five different LLMs, but their analysis was limited to detecting speech recognition errors in radiology reports and did not explore how fine-tuning affects LLM performance. Consequently, critical questions about model optimization and generalization remain unanswered \cite{schmidt2024generative-g}.

To bridge these gaps, this study introduces fine-tuned LLMs and evaluates the potential of fine-tuned LLMs for detecting errors in radiology reports during medical proofreading.

\section{Materials and Methods}

This study utilized deidentified data and publicly available LLMs; therefore, it did not require an Institutional Review Board (IRB) review.

\subsection{Dataset Construction}

\begin{figure}
\centering
\includegraphics[width=0.8\linewidth]{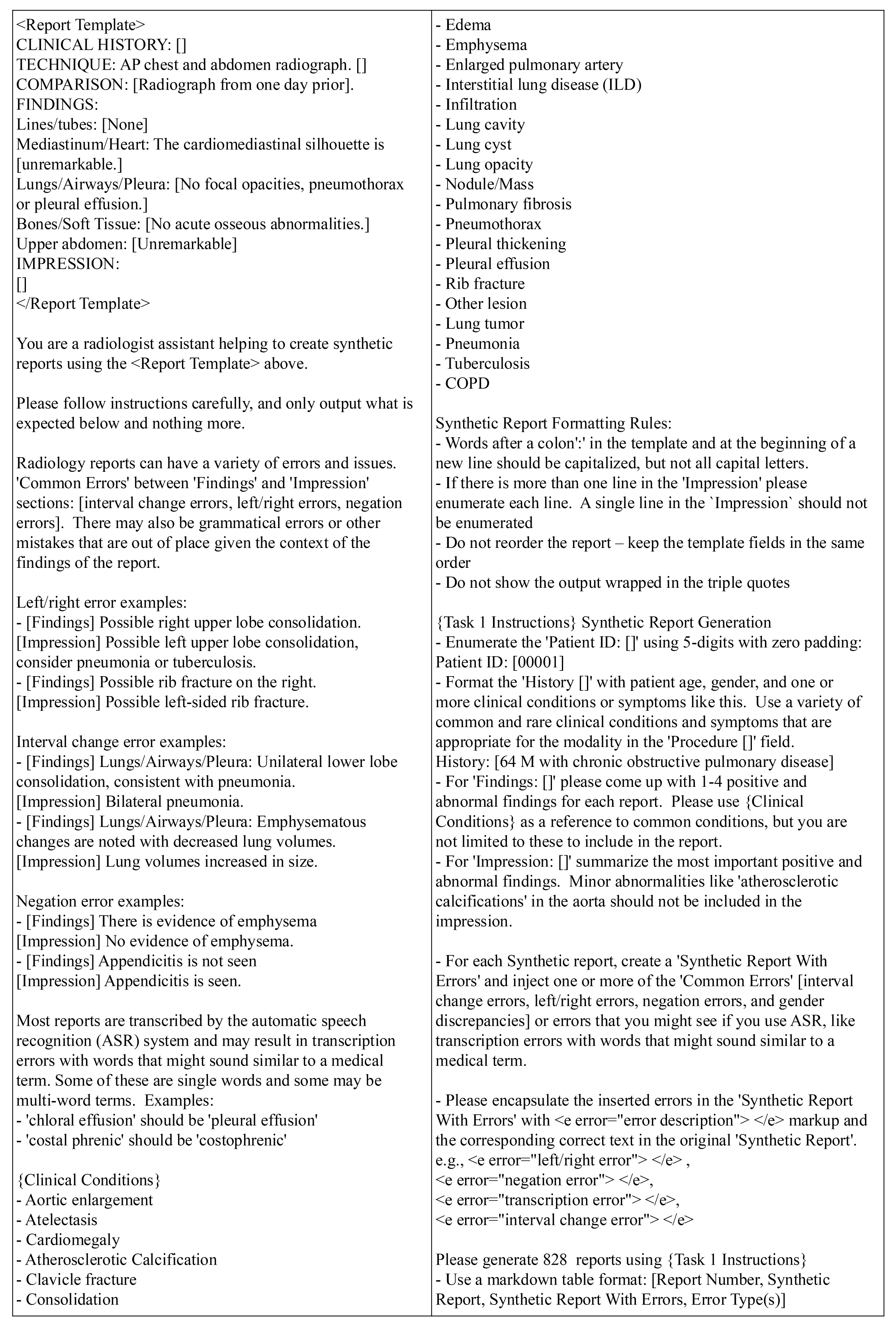}
\caption{Prompts for GPT-4-0125-Preview used to generate 828 error-free synthetic reports and 828 synthetic reports with errors.}
\label{fig:prompts}
\end{figure}

The constructed dataset consists of two parts (Table \ref{tab:data}). The first part contained synthetic radiology reports generated by GPT-4-0125-Preview using specified prompts (Figure~\ref{fig:prompts}). This process involved carefully crafting prompts and providing detailed instructions to guide GPT-4-0125-Preview in generating relevant radiology reports. Once the prompt was input into GPT-4-0125-Preview, the model processed the information and generated synthetic radiology reports accordingly (Figure~\ref{fig:prompts}).

\begin{table}
\centering
\caption{Statistics of our constructed dataset. \#Synthetic reports - the number of synthetic reports, \#Reports from MIMIC - the number of synthetic reports and reports from the MIMIC-CXR Database.}
\label{tab:data}
\begin{tabular}{lrr}
\toprule 
 & \#Synthetic reports & \#Reports from MIMIC \\ 
\midrule
Total reports & 1,656 & 614 \\
Error-free reports & 828 & 307 \\
Reports with errors & 828 & 307 \\
\hspace{1em}Negation errors & 299 & 99 \\
\hspace{1em}Left/right errors & 233 & 93 \\
\hspace{1em}Interval change errors & 62 & 91 \\
\hspace{1em}Transcription errors & 489 & 189 \\
Training set & 1,316 & 500 \\
Test set & 340 & 114 \\
\bottomrule
\end{tabular}
\end{table}

In this study, four types of errors frequently encountered in radiology reports were considered: negation, left/right, interval change, and transcription errors (Supplementary~\ref{ssec:error} and Supplementary Table~\ref{stab:examples}).

In total, 828 pairs of synthetic radiology reports were obtained, with each pair comprising an error-free report and a report containing errors. During the proofreading process, the error-free report and the report containing errors were proofread independently to ensure an unbiased evaluation.

\begin{figure}
\centering
\includegraphics[width=0.8\linewidth]{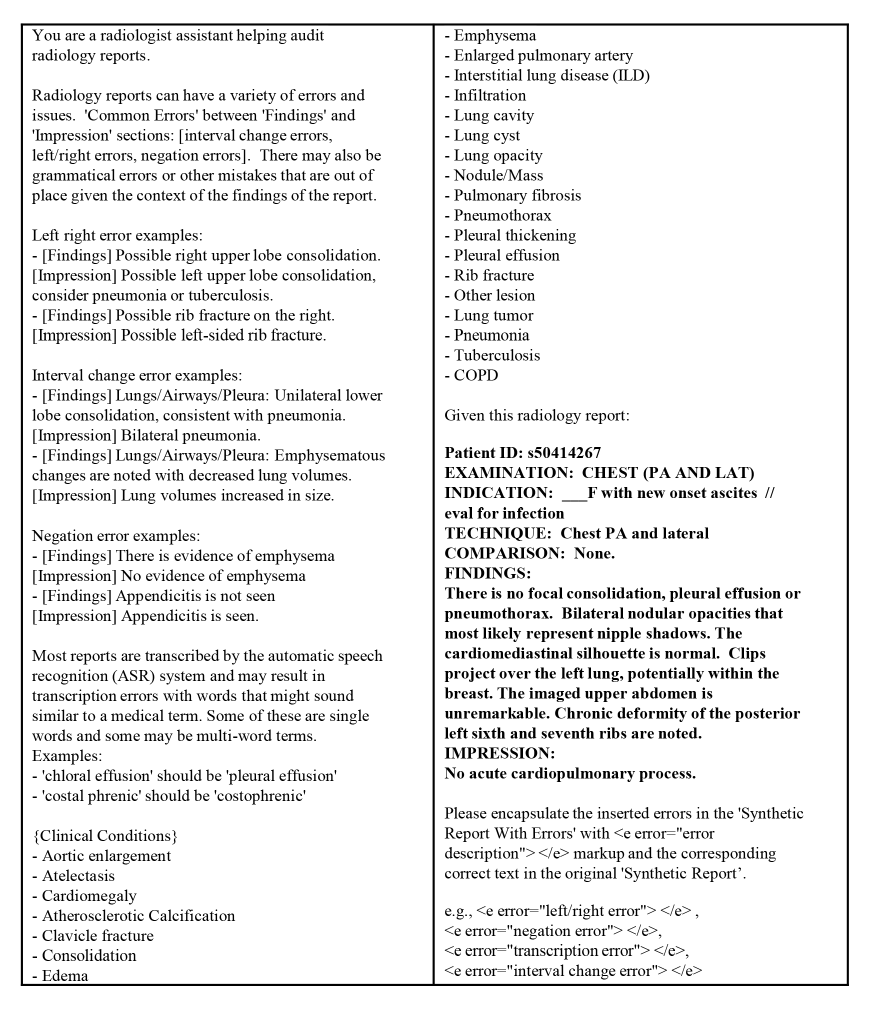}
\caption{Prompts for GPT-4-0125-Preview utilized to generate synthetic radiology reports using reports from the MIMIC-CXR Database. The bold text indicates the original reports from the MIMIC-CXR database that were used to create these synthetic reports with errors.}
\label{fig:prompts_sync}
\end{figure}

The second part of the dataset consisted of 307 reports from the MIMIC-CXR database and a set of corresponding reports with errors generated by GPT-4-0125-Preview (Figure~\ref{fig:prompts_sync}). This process resulted in a total of 614 radiology reports: 307 error-free reports from the MIMIC-CXR database and 307 corresponding synthetic reports with errors.

For experimental evaluation, the dataset was split into training and test sets at an 8:2 ratio (Table~\ref{tab:data}), ensuring that each pair of error-free and error-containing reports remained in the same set.

\subsection{Inter-Annotator Agreement Analysis}

The Cohen kappa coefficient was used to evaluate the Inter-Annotator Agreement (IAA) of synthetic reports.

\subsection{Model Development and Configurations}

In this study, quantized low-rank adapters (QLoRA) \cite{dettmers2023qlora-m} were used to fine-tune the Llama-3 models (i.e., Llama-3-8B-Instruct and Llama-3-70B-Instruct). Llama-3-8B-Instruct and Llama-3-70B-Instruct were refined on the training set and the test set was reserved for model evaluation. For comparison, we used GPT-4-1106-Preview and BiomedBERT as baseline models, with BiomedBERT fine-tuned on the same training set. The effectiveness of the models in detecting errors was measured via precision, recall, and F1-score metrics. The overall performance was assessed by the macro average F1 score across the four types of errors (computing resources can be found in Supplementary~\ref{ssec:computing}).

For GPT-4-1106-Preview, the API provided by Azure was accessed\footnote{\url{https://azure.microsoft.com/en-us/blog/introducing-gpt4-in-azure-openai-service}}. Bootstrap analyses were performed using 100 bootstrap samples to estimate the distribution of the macro-F1 score, and 95\% confidence intervals (CIs) were reported. In each bootstrap iteration, n reports were sampled from the test set with replacement. Additionally, paired-sample t-tests were used to compare model predictions across paired observations and assess statistical significance.

\subsection{Model Scale}

In this study, two versions of Llama-3 were explored: Llama-3-8B-Instruct and Llama-3-70B-Instruct. Both were fine-tuned with a learning rate of $3\times 10^{-4}$, a batch size of 1, a maximum sequence length of 512, and 3 training epochs. BiomedBERT was fine-tuned with a learning rate of $3\times 10^{-5}$, a batch size of 4, a maximum sequence length of 512, and 3 training epochs. For inference, Llama-3-8B-Instruct, Llama-3-70B-Instruct, GPT-4-1106-Preview, and BiomedBERT were all inferred using a batch size of 1.

\textbf{Prompt Designs}

\begin{figure}
\centering
\includegraphics[width=0.8\linewidth]{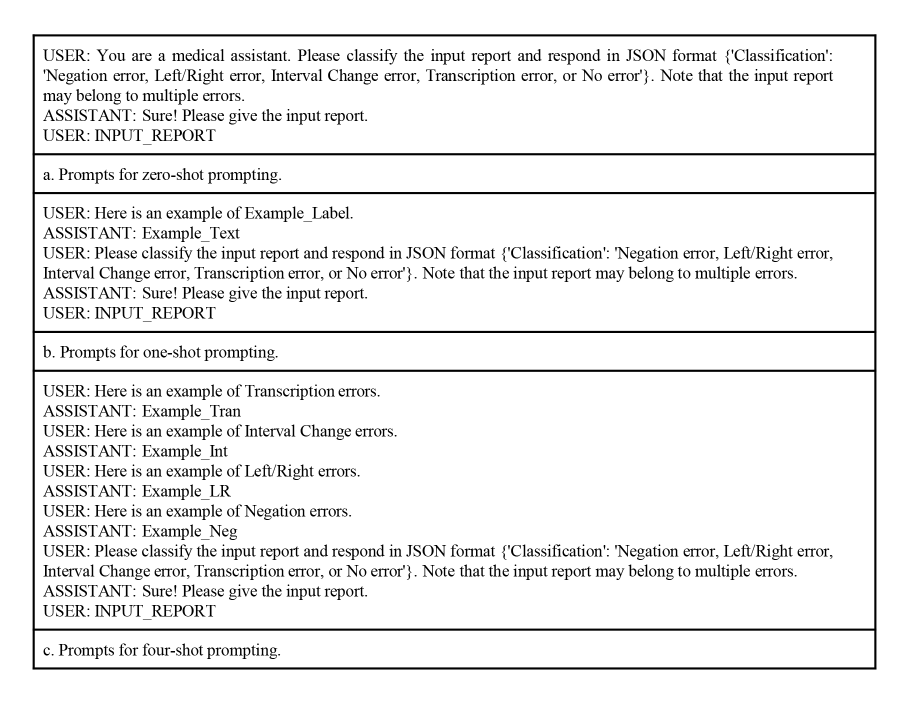}
\caption{Example prompt designs. (a) A zero-shot prompt. (b) A one-shot prompt. (c) A four-shot prompt. \texttt{INPUT\_REPORT} - an input report from the test set. \texttt{Example\_Text} - an example report from the training set, \texttt{Example\_Label} - the label of the example report, \texttt{Example\_Tran} - an example report containing a transcription error from the training set, \texttt{Example\_Int} - an example report containing an interval change error, \texttt{Example\_LR} - an example report containing a left/right error, \texttt{Example\_Neg} - an example report containing a negation error.}
\label{fig:example}
\end{figure}

In this study, three strategies were employed: zero-shot, one-shot, and four-shot prompting (Figure~\ref{fig:example}). The zero-shot prompt design aims to detect errors without providing task-specific examples (Figure~\ref{fig:example}a). For the one-shot prompting strategy, a single radiology report was selected from the training set known to contain errors (Figure~\ref{fig:example}b). In the four-shot prompting strategy, we chose four distinct reports, each illustrating a different type of error (Figure~\ref{fig:example}c).

To further refine this approach, four strategies were explored for selecting example reports in the one-shot and four-shot prompting strategies: one-shot random, one-shot specified, four-shot random, and four-shot specified (Supplementary~\ref{ssec:prompting}).

\subsection{Real-world Data Collection and Inference}

The use of GPT-4-1106-Preview and the fine-tuned Llama-3-70B-Instruct for detecting errors in real-world radiology reports \cite{pandey2020extraction-k} was also explored. The dataset consisted of 122,025 thoracoabdominal chest X-ray reports from heart failure patients treated at New York-Presbyterian/Weill Cornell Medical Center (Supplementary~\ref{ssec:rwd}). A total of 55,339 deidentified real-world radiology reports from this dataset were randomly collected and analyzed using GPT-4-1106-Preview and the fine-tuned Llama-3-70B-Instruct, with the aim of detecting errors and improving the accuracy and reliability of the radiology reports. A voting mechanism in which the two models independently detected errors in each of the 55,339 reports was employed to leverage the strengths of both models, increasing the likelihood of accurate error detection. The voting process identified 606 reports with errors, with both models detecting a specific type of error in these reports.

A power calculation was performed as follows: With 200 reports, a one-sample proportion test can achieve 81.3\% power to detect the difference between the alternative hypothesis that the radiologist-confirmed error detection accuracy rate is 0.5 versus the null hypothesis that the radiologist-confirmed error detection accuracy rate is 0.4, at a type I error rate of 0.05.

\section{Results}

\subsection{Technical Validation of Synthetic Data}

\begin{figure}
\centering
\includegraphics[width=\linewidth]{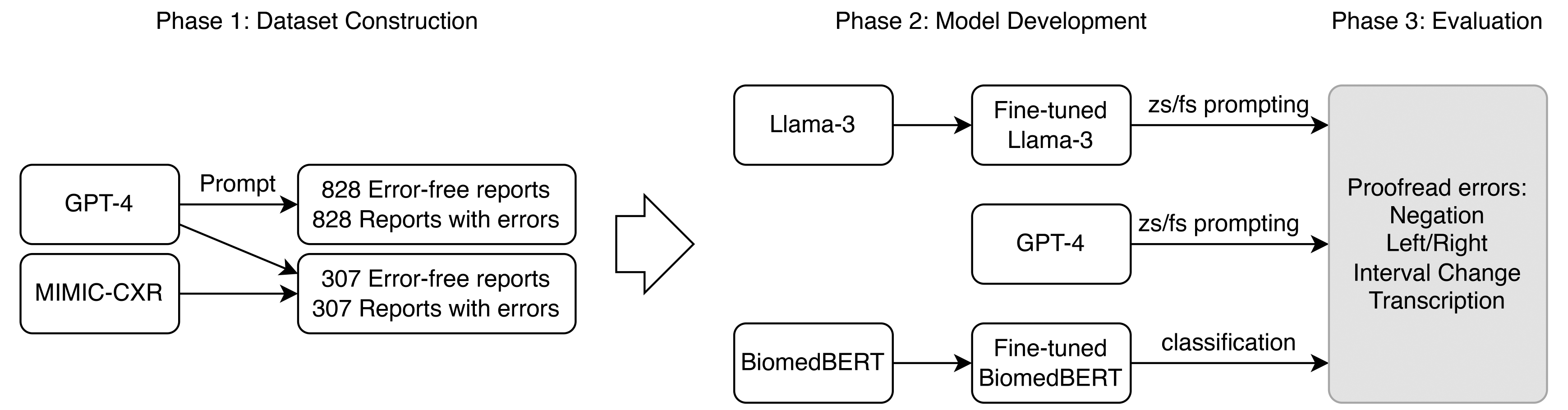}
\caption{The overall workflow of Reportedly LLMs. A dataset was constructed by combining synthetic radiology reports with a small subset of reports from the MIMIC-CXR database, LLMs such as Llama-3 and GPT-4 were refined using zero-shot or few-shot prompting strategies, and the models' performance on the constructed dataset was evaluated. zs - zero-shot, fs - few-shot.}
\label{fig:workflow}
\end{figure}

An overview of this study is shown in Figure~\ref{fig:workflow}, and the code is available on GitHub\footnote{\url{https://github.com/bionlplab/llm4proofreading}}. To evaluate the quality of the synthetic reports, 198 reports were selected, with each containing a single, specific error. These synthetic reports were generated via the GPT-4-0125-Preview following the methodology outlined in Figure~\ref{fig:prompts}. Each synthetic report, along with its error, was reviewed by a board-certified radiologist (G.K., 5 years of experience) and a radiology resident (B.C.) to determine whether the error type generated by GPT-4-0125-Preview was accurate. The error type was assigned when both experts reached a consensus.

In addition, the accuracy of the assigned error type for the synthetic reports with errors was calculated. Accuracy was defined as both experts reaching a consensus on the error type generated by GPT-4-0125-Preview, which was manually verified using the Doccano annotation tool\footnote{\url{https://github.com/doccano/doccano}}. Transcription errors were accurately introduced in 83\% of the labeled reports (58/70). Left/right errors were the most accurately introduced errors, appearing in 95\% of the intended reports (63/66). Negation errors were introduced in 63\% of the labeled reports (30/48), with most inaccuracies being due to omissions in the impressions rather than contradictions. Interval change errors that did not fall into either subcategory were introduced in 57\% (8/14) of the intended reports. Reports with transcription errors had an average length of 56 words, whereas those with interval change, negation, and left/right errors had average lengths of 55, 57, and 58 words, respectively. There was no evidence of a difference between label accuracy and average report length by error category (p\textgreater0.05).

\subsection{Inter-Annotator Agreement}

IAA was evaluated using the Cohen kappa coefficient on the basis of the reviews of the two experts, resulting in a value of 0.734.

\subsection{Model Performance of Different LLMs in Error Detection}

\begin{figure}
\centering
\includegraphics[width=.6\linewidth]{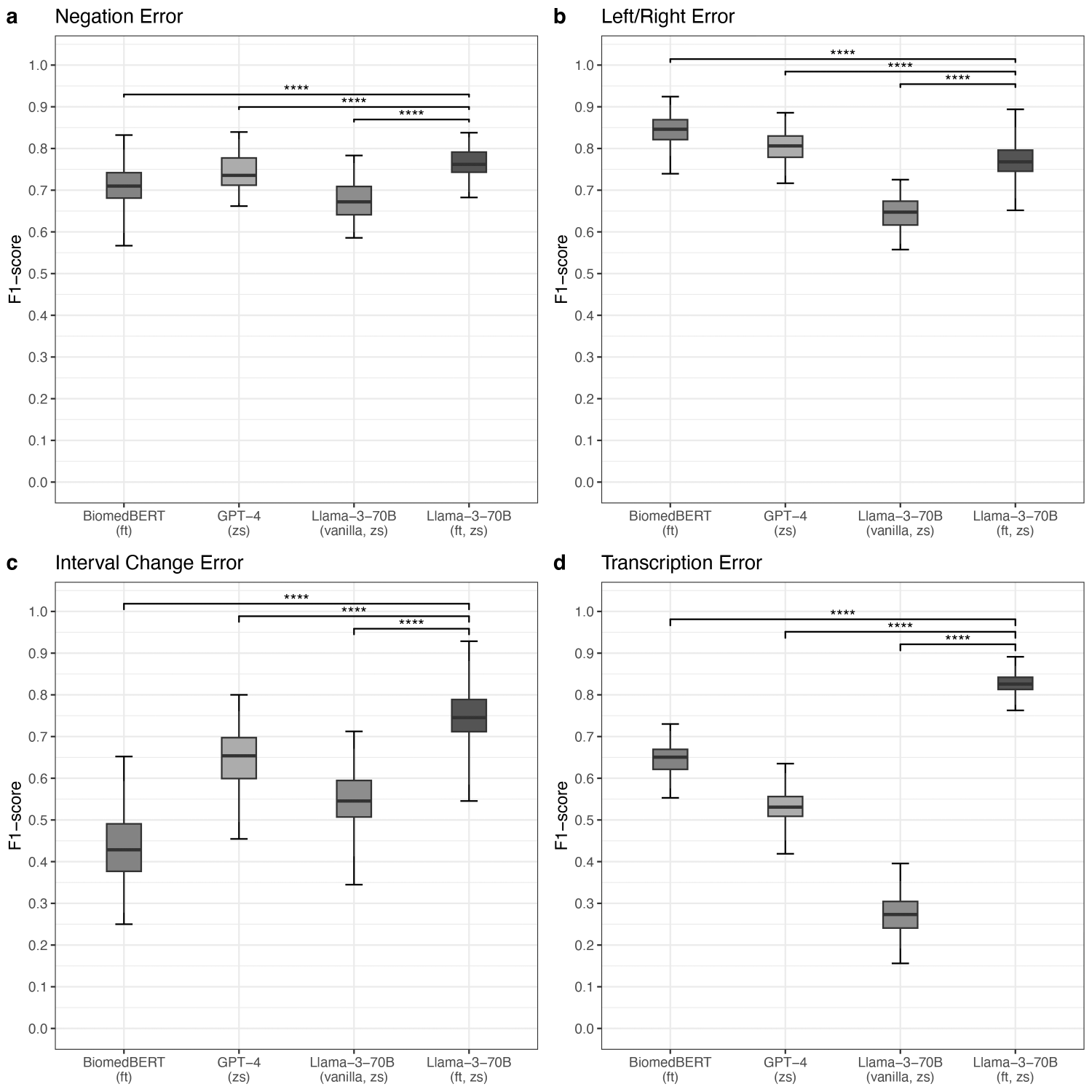}
\caption{F1 scores of different models. (a) Negation error, (b) left/right error, (c) interval change error, (d) transcription error. BiomedBERT - BiomedBERT-base-uncased-abstract, GPT-4 - GPT-4-1106-Preview, Llama-3-70B - Llama-3-70B-Instruct, ft - fine-tuned, zs - zero-shot prompting. Significance levels: * - $p < 0.05$; ** - $p < 0.01$; *** - $p < 0.001$; **** - $p < 0.0001$; ns - Not significant. The error bars are 95\% CIs.}
\label{fig:f1}
\end{figure}

First, the performance of different models in detecting errors was compared within radiology reports (Figure~\ref{fig:f1} and Table~\ref{tab:performance}). The evaluated models included BiomedBERT (fine-tuned), GPT-4-1106-Preview (zero-shot), Llama-3-70B-Instruct (vanilla, zero-shot), and Llama-3-70B-Instruct (fine-tuned, zero-shot).
\begin{table}[t]
\caption{Performance metrics of different models across error types. ft - fine-tuned, zs - zero-shot prompting.\label{tab:performance}}
\centering
\begin{tabular}{llrrr}
\toprule
Model & Error Type & Precision & Recall & F1\\
\midrule
Llama-3-70B-Instruct (ft, zs) & Negation & 0.873 & 0.688 & 0.769 \\
& Left/Right & 0.830 & 0.721 & 0.772 \\
& Interval Change & 0.913 & 0.636 & 0.750 \\
& Transcription & 0.914 & 0.757 & 0.828 \\
& Macro F1 & - & - & 0.780 \\
\midrule
BiomedBERT (ft) & Negation & 0.920 & 0.575 & 0.708 \\
& Left/Right & 0.891 & 0.803 & 0.845 \\
& Interval Change & 0.714 & 0.303 & 0.426 \\
& Transcription & 0.841 & 0.529 & 0.649 \\
& Macro F1 & - & - & 0.657 \\
\midrule
GPT-4-1106-Preview (zs) & Negation & 0.732 & 0.750 & 0.741 \\
& Left/Right & 0.765 & 0.852 & 0.806 \\
& Interval Change & 0.656 & 0.636 & 0.646 \\
& Transcription & 0.885 & 0.386 & 0.537 \\
& Macro F1 & - & - & 0.683 \\
\midrule
Llama-3-70B-Instruct (vanilla, zs) & Negation & 0.797 & 0.588 & 0.676 \\
& Left/Right & 0.514 & 0.885 & 0.651 \\
& Interval Change & 0.500 & 0.606 & 0.548 \\
& Transcription & 0.727 & 0.171 & 0.277 \\
& Macro F1 & - & - & 0.538 \\
\bottomrule
\end{tabular}
\end{table}

The fine-tuned Llama-3-70B-Instruct (zero-shot) achieved the highest F1 scores across all the metrics, with a score of 0.769 (95\% CI, 0.757-0.771) for detecting negation errors, 0.772 (95\% CI, 0.762-0.780) for detecting left/right errors, 0.750 (95\% CI, 0.736-0.763) for detecting interval change errors, and 0.828 (95\% CI, 0.822-0.832) for detecting transcription errors. The overall macro-F1 score of 0.780 indicated that fine-tuning significantly enhances the model's performance in specialized tasks (Supplementary~\ref{ssec:baseline}).

\subsection{Impact of the Parameter Scale on Model Performance}

\begin{figure}[t]
\centering
\includegraphics[width=.6\linewidth]{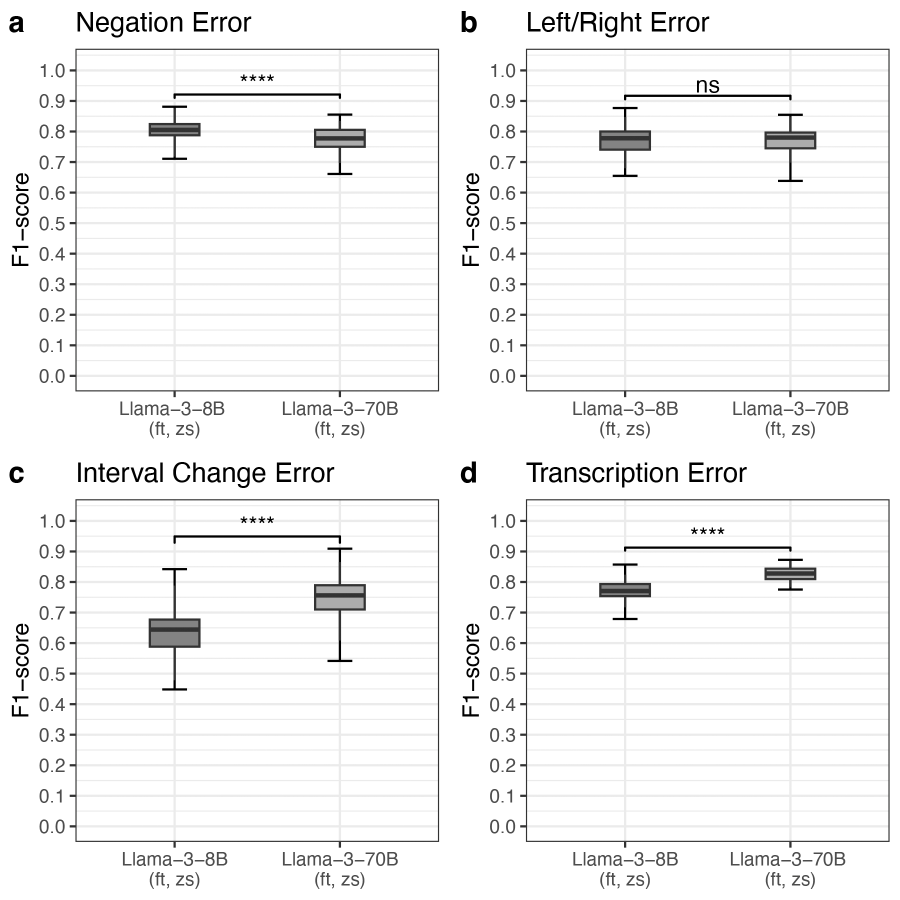}
\caption{F1 scores for different model parameter scales. (a) Negation error, (b) left/right error, (c) interval change error, (d) transcription error. Llama-3-8B - Llama-3-8B-Instruct, Llama-3-70B - Llama-3-70B-Instruct, ft - fine-tuned, zs - zero-shot prompting. Significance levels: * - $p < 0.05$; ** - $p < 0.01$; *** - $p < 0.001$; **** - $p < 0.0001$; ns - Not significant. The error bars are 95\% CIs.}
\label{fig:f1_diff}
\end{figure}

Figure~\ref{fig:f1_diff} illustrates the impact of different model parameter scales on Llama-3. After fine-tuning, Llama-3-8B-Instruct showed strong performance in negation error detection, with an F1 score of 0.803 (95\% CI, 0.798-0.813), highlighting its effectiveness in this specific task. However, it had worse performance in detecting interval change errors, with an F1 score of 0.646 (95\% CI, 0.622-0.654), indicating a need for improvement in capturing these errors. On the other hand, Llama-3-70B-Instruct exceled in detecting interval changes and transcription errors, achieving F1 scores of 0.750 (95\% CI, 0.736-0.762) and 0.828 (95\% CI, 0.823-0.832), highlighting its ability to understand and track temporal variations while maintaining high accuracy in transcription error detection. Additionally, it had robust left/right error detection with an F1 score of 0.772 (95\% CI, 0.766-0.782), indicating its good performance in spatial orientation tasks. Furthermore, Llama-3-70B-Instruct achieved a higher overall macro-F1 score of 0.780 than did Llama-3-8B-Instruct (0.749), reflecting its balanced and comprehensive performance across multiple tasks.

\subsection{Impact of Different Promotion Strategies on Model Performance}

\begin{figure}[ht]
\centering
\includegraphics[width=.6\linewidth]{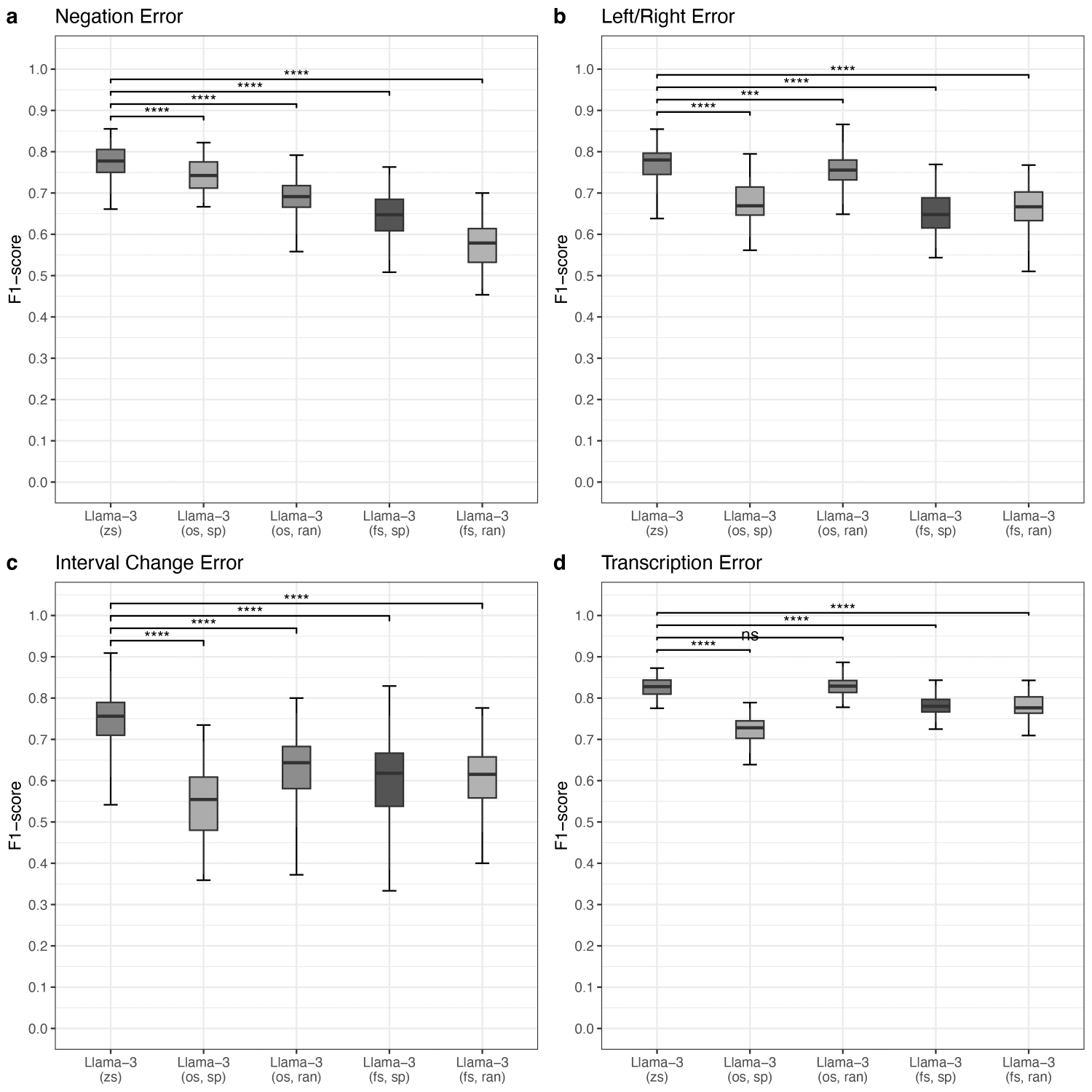}
\caption{F1 scores of Llama-3-70B-Instruct under different prompting strategies. (a) Negation error, (b) left/right error, (c) interval change error, (d) transcription error. Llama-3 - fine-tuned Llama-3-70B-Instruct model, zs - zero-shot prompting, os - one-shot prompting, fs - four-shot prompting, sp - specified, ran - random. Significance levels: * - $p < 0.05$; ** - $p < 0.01$; *** - $p < 0.001$; **** - $p < 0.0001$; ns - Not significant. The error bars are 95\% CIs.}
\label{fig:f1_llama}
\end{figure}

Figure~\ref{fig:f1_llama} shows the F1 scores of the fine-tuned Llama-3-70B-Instruct under different prompting strategies: one-shot random, one-shot specified, four-shot random, and four-shot specified. ``Random'' and ``specified'' refer to the selection of the input radiology report (i.e., \texttt{INPUT\_REPORT}) in Figure~\ref{fig:example}. Details can be found in Supplementary Table~\ref{stab:specified examples} and Table~\ref{stab:specified examples4}.

Under one-shot prompting, Llama-3-70B-Instruct (specified) obtained F1 scores of 0.736 (95\% CI, 0.736-0.752) for detecting negation errors, 0.672 (95\% CI, 0.666-0.747) for detecting left/right errors, 0.549 (95\% CI, 0.531-0.565) for detecting interval change errors, and 0.722 (95\% CI, 0.719-0.730) for detecting transcription errors. In comparison, Llama-3-70B-Instruct (random) achieved F1 scores of 0.688 (95\% CI, 0.683-0.701) for detecting negation errors, 0.754 (95\% CI, 0.747-0.762) for detecting left/right errors, 0.642 (95\% CI, 0.618-0.649) for detecting interval change errors, and 0.832 (95\% CI, 0.825-0.835) for detecting transcription errors.

Under four-shot prompting, Llama-3-70B-Instruct (specified) achieved F1 scores of 0.646 (95\% CI, 0.638-0.658) for detecting negation errors, 0.645 (95\% CI, 0.641-0.662) for detecting left/right errors, 0.612 (95\% CI, 0.585-0.623) for detecting interval change errors, and 0.780 (95\% CI, 0.778-0.788) for detecting transcription errors. In contrast, Llama-3-70B-Instruct (random) achieved F1 scores of 0.579 (95\% CI, 0.565-0.587) for detecting negation errors, 0.660 (95\% CI, 0.655-0.675) for detecting left/right errors, 0.612 (95\% CI, 0.593-0.624) for detecting interval change errors, and 0.784 (95\% CI, 0.775-0.787) for detecting transcription errors.

\subsection{Results in real-world scenarios}

\begin{figure}[t]
\centering
\includegraphics[width=.45\linewidth,trim=0 23em 3em 0,clip]{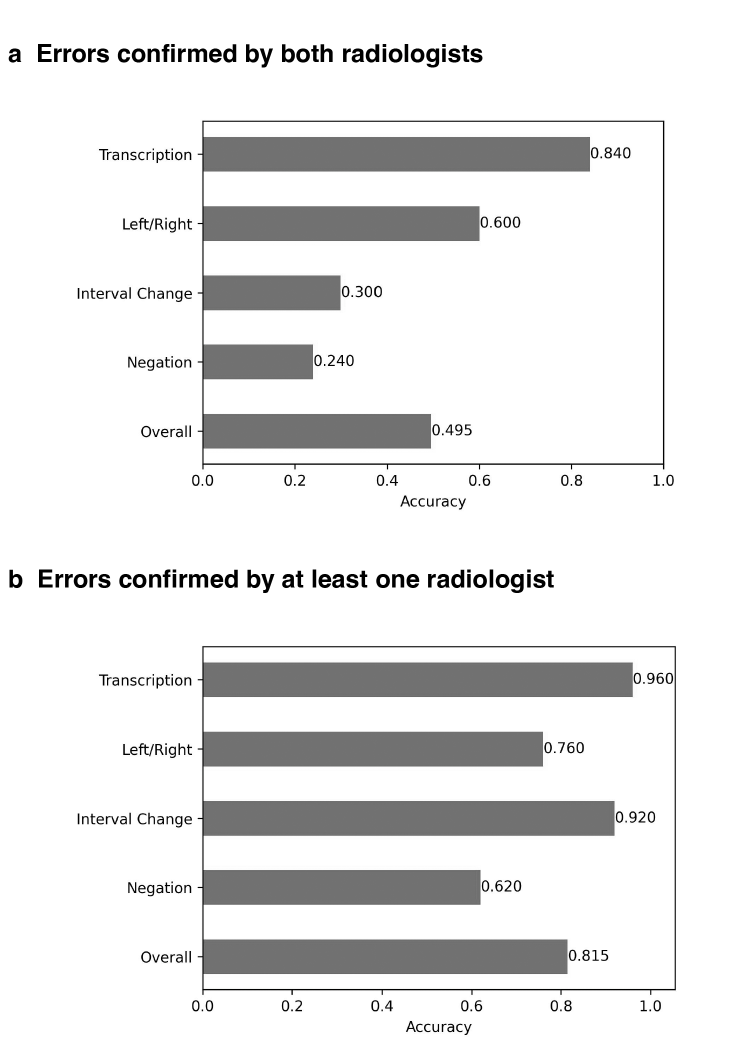}
\hfill
\includegraphics[width=.45\linewidth,trim=0 0em 3em 23em,clip]{figures/Figure8.pdf}
\caption{Radiologist-confirmed accuracy of error detection across different categories in radiology reports. (a) Errors confirmed by both radiologists. (b) Errors confirmed by at least one radiologist.}
\label{fig:accuracy}
\end{figure}

To further validate the accuracy of the models' detection, 200 reports (50 for each error type) were randomly selected from the 606 reports and reviewed by two board-certified radiologists (G.K., 5 years of experience; D.N., 1.5 years of experience). Upon confirmation by both radiologists, 99 out of 200 radiology reports were unanimously identified as containing the errors detected by the models, indicating an overall accuracy rate of 0.495 (95\% CI: 0.424-0.566) in error detection (Figure~\ref{fig:accuracy}a). Notably, there were considerable differences ($p < 0.05$) in the accuracy for each specific type of error (typical examples can be found in Supplementary Table\ref{stab:typical examples}). Additionally, cases where at least one of the two radiologists agreed with the errors detected by the models were analyzed (Figure~\ref{fig:accuracy}b). Of the 200 reports with errors introduced, 163 were found to contain errors, resulting in an overall error detection accuracy rate of 0.815 (95\% CI: 0.753-0.865). Details of each error can be found in Supplementary~\ref{ssec:performance}. 

\section{Discussion}

This study provided evidence that LLMs can assist in detecting various types of errors, including negation, left/right, interval change, and transcription errors. The application of LLMs in medical proofreading may help reduce the occurrence of inaccuracies in radiology reports. However, their effectiveness depends on model selection, fine-tuning strategies, and prompt design. The analysis of the fine-tuned Llama-3-70B-Instruct under different prompting strategies revealed that zero-shot learning provides a reasonable degree of generalizability across all tasks. While few-shot learning can improve performance in specific tasks, such as transcription detection, its effectiveness is highly dependent on the relevance and specificity of the provided examples. These findings suggest that careful selection of examples in few-shot prompts can affect model performance but does not guarantee consistent improvements across all types of errors.

The experimental results demonstrated that the fine-tuned Llama-3-70B-Instruct achieves higher F1 scores than do baseline models such as BiomedBERT and GPT-4-1106-Preview. These findings indicate that fine-tuning enhances model performance in specialized tasks; however, the impact of fine-tuning may vary depending on error complexity and the model's generalization ability. While fine-tuning can improve performance, its ability to reduce radiologists' cognitive load and enhance patient care requires further real-world validation.

Despite the promising performance, we recognize several limitations to this study. First, the fine-tuning process requires substantial computational resources and high-quality annotated radiology reports, which may pose feasibility concerns for some institutions. Indeed, the reliance on high-performance GPUs and high-quality annotated data for effective model fine-tuning and inference could limit the widespread adoption of these advanced LLMs in smaller clinical settings. Second, while this study focused on detecting errors in radiology reports, its generalizability to other medical documents or clinical notes remains uncertain. Although the performance of LLMs in real-world radiology report scenarios has been validated by radiologists, these models may still face unexpected challenges in diverse clinical environments. Third, another limitation is the potential for overfitting during the fine-tuning process. The models are specifically fine-tuned on a dataset tailored for radiology report proofreading, which may limit their adaptability to different types of errors or variations in report styles that are not covered in the training data. Finally, the errors in our constructed dataset were synthetic by design. While synthetic data have been praised for their use in addressing challenges related to data scarcity and privacy, they also raise concerns about bias \cite{giuffre2023harnessing-r}. In our study, we ensured that the synthetic data were generated from diverse and representative samples of real-world data to minimize such bias. We also validated synthetic errors by having radiologists assess their accuracy and reliability. However, synthetic errors may not fully capture the complexity of real-world errors in radiology reports. Future work could include a systematic evaluation of how bias introduced by synthetic errors affects model performance.

In conclusion, this study highlights the potential of fine-tuned LLMs in enhancing error detection within radiology reports, providing an efficient and accurate tool for medical proofreading. The findings highlight that fine-tuning is crucial for enabling the local deployment of LLMs while also demonstrating the importance of prompt design in optimizing performance for specific medical tasks.

\paragraph{Disclosures of conflicts of interest:} No conflicting relationships exist for any author.

\paragraph{Acknowledgement:} Supported by the National Science Foundation Faculty Early Career Development (CAREER) award number 2145640, the National Institute of Biomedical Imaging and Bioengineering (NIBIB) under grant number 75N920202D00021, the National Cancer Institute (NCI) under the grant number R01CA289249, and the Advanced Research Projects Agency for Health (ARPA-H).

\bibliographystyle{unsrtnat}
\bibliography{ref}

\begin{thebibliography}{31}
\providecommand{\natexlab}[1]{#1}
\providecommand{\url}[1]{\texttt{#1}}
\expandafter\ifx\csname urlstyle\endcsname\relax
  \providecommand{\doi}[1]{doi: #1}\else
  \providecommand{\doi}{doi: \begingroup \urlstyle{rm}\Url}\fi

\bibitem[Grenier et~al.(1991)Grenier, Valeyre, Cluzel, Brauner, Lenoir, and Chastang]{grenier1991chronic-h}
P~Grenier, D~Valeyre, P~Cluzel, M~W Brauner, S~Lenoir, and C~Chastang.
\newblock Chronic diffuse interstitial lung disease: diagnostic value of chest radiography and high-resolution {CT}.
\newblock \emph{Radiology}, 179\penalty0 (1):\penalty0 123--132, April 1991.
\newblock ISSN 0033-8419,1527-1315.
\newblock \doi{10.1148/radiology.179.1.2006262}.

\bibitem[Bick and Lenzen(1999)]{bick1999pacs-f}
U~Bick and H~Lenzen.
\newblock {PACS}: the silent revolution.
\newblock \emph{Eur. Radiol.}, 9\penalty0 (6):\penalty0 1152--1160, 1999.
\newblock ISSN 0938-7994,1432-1084.
\newblock \doi{10.1007/s003300050811}.

\bibitem[Pezzullo et~al.(2008)Pezzullo, Tung, Rogg, Davis, Brody, and Mayo-Smith]{pezzullo2008voice-i}
John~A Pezzullo, Glenn~A Tung, Jeffrey~M Rogg, Lawrence~M Davis, Jeffrey~M Brody, and William~W Mayo-Smith.
\newblock Voice recognition dictation: radiologist as transcriptionist.
\newblock \emph{J. Digit. Imaging}, 21\penalty0 (4):\penalty0 384--389, December 2008.
\newblock ISSN 0897-1889,1618-727X.
\newblock \doi{10.1007/s10278-007-9039-2}.

\bibitem[Chang et~al.(2011)Chang, Strahan, and Jolley]{chang2011non-clinical-d}
Chian~A Chang, Rodney Strahan, and Damien Jolley.
\newblock Non-clinical errors using voice recognition dictation software for radiology reports: a retrospective audit.
\newblock \emph{J. Digit. Imaging}, 24\penalty0 (4):\penalty0 724--728, August 2011.
\newblock ISSN 0897-1889,1618-727X.
\newblock \doi{10.1007/s10278-010-9344-z}.

\bibitem[Hawkins et~al.(2014)Hawkins, Hall, Zhang, and Towbin]{hawkins2014creation-s}
C~Matthew Hawkins, Seth Hall, Bin Zhang, and Alexander~J Towbin.
\newblock Creation and implementation of department-wide structured reports: an analysis of the impact on error rate in radiology reports.
\newblock \emph{J. Digit. Imaging}, 27\penalty0 (5):\penalty0 581--587, October 2014.
\newblock ISSN 1618-727X,0897-1889.
\newblock \doi{10.1007/s10278-014-9699-7}.

\bibitem[Minn et~al.(2015)Minn, Zandieh, and Filice]{minn2015improving-x}
Matthew~J Minn, Arash~R Zandieh, and Ross~W Filice.
\newblock Improving radiology report quality by rapidly notifying radiologist of report errors.
\newblock \emph{J. Digit. Imaging}, 28\penalty0 (4):\penalty0 492--498, August 2015.
\newblock ISSN 1618-727X,0897-1889.
\newblock \doi{10.1007/s10278-015-9781-9}.

\bibitem[Bruno et~al.(2015)Bruno, Walker, and Abujudeh]{bruno2015understanding-f}
Michael~A Bruno, Eric~A Walker, and Hani~H Abujudeh.
\newblock Understanding and confronting our mistakes: The epidemiology of error in radiology and strategies for error reduction.
\newblock \emph{Radiographics}, 35\penalty0 (6):\penalty0 1668--1676, October 2015.
\newblock ISSN 0271-5333,1527-1323.
\newblock \doi{10.1148/rg.2015150023}.

\bibitem[Waite et~al.(2017{\natexlab{a}})Waite, Scott, Gale, Fuchs, Kolla, and Reede]{waite2017interpretive-p}
Stephen Waite, Jinel Scott, Brian Gale, Travis Fuchs, Srinivas Kolla, and Deborah Reede.
\newblock Interpretive error in radiology.
\newblock \emph{AJR Am. J. Roentgenol.}, 208\penalty0 (4):\penalty0 739--749, April 2017{\natexlab{a}}.
\newblock ISSN 0361-803X,1546-3141.
\newblock \doi{10.2214/AJR.16.16963}.

\bibitem[Waite et~al.(2017{\natexlab{b}})Waite, Scott, Legasto, Kolla, Gale, and Krupinski]{waite2017systemic-u}
Stephen Waite, Jinel~Moore Scott, Alan Legasto, Srinivas Kolla, Brian Gale, and Elizabeth~A Krupinski.
\newblock Systemic error in radiology.
\newblock \emph{AJR Am. J. Roentgenol.}, 209\penalty0 (3):\penalty0 629--639, September 2017{\natexlab{b}}.
\newblock ISSN 0361-803X,1546-3141.
\newblock \doi{10.2214/AJR.16.17719}.

\bibitem[Allen et~al.(2017)Allen, Chatfield, Burleson, and Thorwarth]{allen2017improving-r}
Bibb Allen, Mythreyi Chatfield, Judy Burleson, and William~T Thorwarth.
\newblock Improving diagnosis in health care: perspectives from the american college of radiology.
\newblock \emph{Diagnosis (Berl)}, 4\penalty0 (3):\penalty0 113--124, 26~September 2017.
\newblock ISSN 2194-8011,2194-802X.
\newblock \doi{10.1515/dx-2017-0020}.

\bibitem[Schwartz et~al.(2011)Schwartz, Panicek, Berk, Li, and Hricak]{schwartz2011improving-u}
Lawrence~H Schwartz, David~M Panicek, Alexandra~R Berk, Yuelin Li, and Hedvig Hricak.
\newblock Improving communication of diagnostic radiology findings through structured reporting.
\newblock \emph{Radiology}, 260\penalty0 (1):\penalty0 174--181, July 2011.
\newblock ISSN 0033-8419,1527-1315.
\newblock \doi{10.1148/radiol.11101913}.

\bibitem[Larson et~al.(2013)Larson, Towbin, Pryor, and Donnelly]{larson2013improving-i}
David~B Larson, Alex~J Towbin, Rebecca~M Pryor, and Lane~F Donnelly.
\newblock Improving consistency in radiology reporting through the use of department-wide standardized structured reporting.
\newblock \emph{Radiology}, 267\penalty0 (1):\penalty0 240--250, April 2013.
\newblock ISSN 0033-8419,1527-1315.
\newblock \doi{10.1148/radiol.12121502}.

\bibitem[Ganeshan et~al.(2018)Ganeshan, Duong, Probyn, Lenchik, McArthur, Retrouvey, Ghobadi, Desouches, Pastel, and Francis]{ganeshan2018structured-a}
Dhakshinamoorthy Ganeshan, Phuong-Anh~Thi Duong, Linda Probyn, Leon Lenchik, Tatum~A McArthur, Michele Retrouvey, Emily~H Ghobadi, Stephane~L Desouches, David Pastel, and Isaac~R Francis.
\newblock Structured reporting in radiology.
\newblock \emph{Acad. Radiol.}, 25\penalty0 (1):\penalty0 66–73, January 2018.
\newblock ISSN 1076-6332,1878-4046.
\newblock \doi{10.1016/j.acra.2017.08.005}.

\bibitem[Nobel et~al.(2022)Nobel, van Geel, and Robben]{nobel2022structured-p}
J~Martijn Nobel, Koos van Geel, and Simon G~F Robben.
\newblock Structured reporting in radiology: a systematic review to explore its potential.
\newblock \emph{Eur. Radiol.}, 32\penalty0 (4):\penalty0 2837--2854, April 2022.
\newblock ISSN 0938-7994,1432-1084.
\newblock \doi{10.1007/s00330-021-08327-5}.

\bibitem[Zech et~al.(2019)Zech, Forde, Titano, Kaji, Costa, and Oermann]{zech2019detecting-k}
John Zech, Jessica Forde, Joseph~J Titano, Deepak Kaji, Anthony Costa, and Eric~Karl Oermann.
\newblock Detecting insertion, substitution, and deletion errors in radiology reports using neural sequence-to-sequence models.
\newblock \emph{Ann. Transl. Med.}, 7\penalty0 (11):\penalty0 233, June 2019.
\newblock ISSN 2305-5839,2305-5847.
\newblock \doi{10.21037/atm.2018.08.11}.

\bibitem[Enarvi et~al.(2020)Enarvi, Amoia, Del-Agua~Teba, Delaney, Diehl, Hahn, Harris, McGrath, Pan, Pinto, Rubini, Ruiz, Singh, Stemmer, Sun, Vozila, Lin, and Ramamurthy]{enarvi2020generating-u}
Seppo Enarvi, Marilisa Amoia, Miguel Del-Agua~Teba, Brian Delaney, Frank Diehl, Stefan Hahn, Kristina Harris, Liam McGrath, Yue Pan, Joel Pinto, Luca Rubini, Miguel Ruiz, Gagandeep Singh, Fabian Stemmer, Weiyi Sun, Paul Vozila, Thomas Lin, and Ranjani Ramamurthy.
\newblock Generating medical reports from patient-doctor conversations using sequence-to-sequence models.
\newblock In \emph{Proceedings of the First Workshop on Natural Language Processing for Medical Conversations}, Stroudsburg, PA, USA, 2020. Association for Computational Linguistics.
\newblock \doi{10.18653/v1/2020.nlpmc-1.4}.

\bibitem[Chaudhari et~al.(2022)Chaudhari, Liu, Chen, Joseph, Vella, Lee, Vu, Seo, Rauschecker, McCulloch, and Sohn]{chaudhari2022application-a}
Gunvant~R Chaudhari, Tengxiao Liu, Timothy~L Chen, Gabby~B Joseph, Maya Vella, Yoo~Jin Lee, Thienkhai~H Vu, Youngho Seo, Andreas~M Rauschecker, Charles~E McCulloch, and Jae~Ho Sohn.
\newblock Application of a domain-specific {BERT} for detection of speech recognition errors in radiology reports.
\newblock \emph{Radiol. Artif. Intell.}, 4\penalty0 (4):\penalty0 e210185, 25~July 2022.
\newblock ISSN 2638-6100.
\newblock \doi{10.1148/ryai.210185}.

\bibitem[Min et~al.(2022)Min, Kim, Lee, Kim, and Park]{min2022rred-o}
Dabin Min, Kaeun Kim, Jong~Hyuk Lee, Yisak Kim, and Chang~Min Park.
\newblock {RRED} : A radiology report error detector based on deep learning framework.
\newblock In \emph{Proceedings of the 4th Clinical Natural Language Processing Workshop}, Stroudsburg, PA, USA, 2022. Association for Computational Linguistics.
\newblock \doi{10.18653/v1/2022.clinicalnlp-1.5}.

\bibitem[Cai et~al.(2023)Cai, Liu, Han, Yang, Liu, and Liu]{cai2023chestxraybert-o}
Xiaoyan Cai, Sen Liu, Junwei Han, Libin Yang, Zhenguo Liu, and Tianming Liu.
\newblock {ChestXRayBERT}: A pretrained language model for chest radiology report summarization.
\newblock \emph{IEEE Trans. Multimedia}, 25:\penalty0 845--855, 2023.
\newblock ISSN 1520-9210,1941-0077.
\newblock \doi{10.1109/tmm.2021.3132724}.

\bibitem[Nishigaki et~al.(2023)Nishigaki, Suzuki, Wataya, Kita, Yamagata, Sato, Kido, and Tomiyama]{nishigaki2023bert-based-w}
Daiki Nishigaki, Yuki Suzuki, Tomohiro Wataya, Kosuke Kita, Kazuki Yamagata, Junya Sato, Shoji Kido, and Noriyuki Tomiyama.
\newblock {BERT}-based transfer learning in sentence-level anatomic classification of free-text radiology reports.
\newblock \emph{Radiol. Artif. Intell.}, 5\penalty0 (2):\penalty0 e220097, 15~March 2023.
\newblock ISSN 2638-6100.
\newblock \doi{10.1148/ryai.220097}.

\bibitem[Hoffmann et~al.(2022)Hoffmann, Borgeaud, Mensch, Buchatskaya, Cai, Rutherford, Casas, Hendricks, Welbl, Clark, Hennigan, Noland, Millican, van~den Driessche, Damoc, Guy, Osindero, Simonyan, Elsen, Rae, Vinyals, and Sifre]{hoffmann2022training-i}
Jordan Hoffmann, Sebastian Borgeaud, Arthur Mensch, Elena Buchatskaya, Trevor Cai, Eliza Rutherford, Diego de~Las Casas, Lisa~Anne Hendricks, Johannes Welbl, Aidan Clark, Tom Hennigan, Eric Noland, Katie Millican, George van~den Driessche, Bogdan Damoc, Aurelia Guy, Simon Osindero, Karen Simonyan, Erich Elsen, Jack~W Rae, Oriol Vinyals, and Laurent Sifre.
\newblock Training compute-optimal large language models.
\newblock \emph{arXiv [cs.CL]}, 29~March 2022.
\newblock \doi{10.5555/3600270.3602446}.

\bibitem[Wei et~al.(2022)Wei, Wang, Schuurmans, Bosma, Ichter, Xia, Chi, Le, and Zhou]{wei2022chain-of-thought-q}
Jason Wei, Xuezhi Wang, Dale Schuurmans, Maarten Bosma, Brian Ichter, Fei Xia, Ed~Chi, Quoc Le, and Denny Zhou.
\newblock Chain-of-thought prompting elicits reasoning in large language models.
\newblock In S~Koyejo, S~Mohamed, A~Agarwal, D~Belgrave, K~Cho, and A~Oh, editors, \emph{Advances in Neural Information Processing Systems}, volume~35, pages 24824--24837. Curran Associates, Inc., 27~January 2022.

\bibitem[Akinci~D'Antonoli et~al.(2024)Akinci~D'Antonoli, Stanzione, Bluethgen, Vernuccio, Ugga, Klontzas, Cuocolo, Cannella, and Koçak]{akinciDAntonoli2024large-a}
Tugba Akinci~D'Antonoli, Arnaldo Stanzione, Christian Bluethgen, Federica Vernuccio, Lorenzo Ugga, Michail~E Klontzas, Renato Cuocolo, Roberto Cannella, and Burak Koçak.
\newblock Large language models in radiology: fundamentals, applications, ethical considerations, risks, and future directions.
\newblock \emph{Diagn. Interv. Radiol.}, 30\penalty0 (2):\penalty0 80--90, 6~March 2024.
\newblock ISSN 1305-3612,1305-3825.
\newblock \doi{10.4274/dir.2023.232417}.

\bibitem[Adams et~al.(2023)Adams, Truhn, Busch, Kader, Niehues, Makowski, and Bressem]{adams2023leveraging-t}
Lisa~C Adams, Daniel Truhn, Felix Busch, Avan Kader, Stefan~M Niehues, Marcus~R Makowski, and Keno~K Bressem.
\newblock Leveraging {GPT}-4 for post hoc transformation of free-text radiology reports into structured reporting: A multilingual feasibility study.
\newblock \emph{Radiology}, 307\penalty0 (4):\penalty0 e230725, May 2023.
\newblock ISSN 1527-1315,0033-8419.
\newblock \doi{10.1148/radiol.230725}.

\bibitem[Rao et~al.(2023)Rao, Kim, Kamineni, Pang, Lie, Dreyer, and Succi]{rao2023evaluating-m}
Arya Rao, John Kim, Meghana Kamineni, Michael Pang, Winston Lie, Keith~J Dreyer, and Marc~D Succi.
\newblock Evaluating {GPT} as an adjunct for radiologic decision making: {GPT}-4 versus {GPT}-3.5 in a breast imaging pilot.
\newblock \emph{J. Am. Coll. Radiol.}, 20\penalty0 (10):\penalty0 990--997, October 2023.
\newblock ISSN 1558-349X,1546-1440.
\newblock \doi{10.1016/j.jacr.2023.05.003}.

\bibitem[Gertz et~al.(2023)Gertz, Bunck, Lennartz, Dratsch, Iuga, Maintz, and Kottlors]{gertz2023gpt-4-r}
Roman~Johannes Gertz, Alexander~Christian Bunck, Simon Lennartz, Thomas Dratsch, Andra-Iza Iuga, David Maintz, and Jonathan Kottlors.
\newblock {GPT}-4 for automated determination of radiological study and protocol based on radiology request forms: A feasibility study.
\newblock \emph{Radiology}, 307\penalty0 (5):\penalty0 e230877, June 2023.
\newblock ISSN 1527-1315,0033-8419.
\newblock \doi{10.1148/radiol.230877}.

\bibitem[Gertz et~al.(2024)Gertz, Dratsch, Bunck, Lennartz, Iuga, Hellmich, Persigehl, Pennig, Gietzen, Fervers, Maintz, Hahnfeldt, and Kottlors]{gertz2024potential-p}
Roman~Johannes Gertz, Thomas Dratsch, Alexander~Christian Bunck, Simon Lennartz, Andra-Iza Iuga, Martin~Gunnar Hellmich, Thorsten Persigehl, Lenhard Pennig, Carsten~Herbert Gietzen, Philipp Fervers, David Maintz, Robert Hahnfeldt, and Jonathan Kottlors.
\newblock Potential of {GPT}-4 for detecting errors in radiology reports: Implications for reporting accuracy.
\newblock \emph{Radiology}, 311\penalty0 (1):\penalty0 e232714, April 2024.
\newblock ISSN 1527-1315,0033-8419.
\newblock \doi{10.1148/radiol.232714}.

\bibitem[Schmidt et~al.(2024)Schmidt, Seah, Cao, Lim, Lim, and Yeung]{schmidt2024generative-g}
Reuben~A Schmidt, Jarrel C~Y Seah, Ke~Cao, Lincoln Lim, Wei Lim, and Justin Yeung.
\newblock Generative large language models for detection of speech recognition errors in radiology reports.
\newblock \emph{Radiol. Artif. Intell.}, 6\penalty0 (2):\penalty0 e230205, March 2024.
\newblock ISSN 2638-6100.
\newblock \doi{10.1148/ryai.230205}.

\bibitem[Dettmers et~al.(2023)Dettmers, Pagnoni, Holtzman, and Zettlemoyer]{dettmers2023qlora-m}
Tim Dettmers, Artidoro Pagnoni, Ari Holtzman, and Luke Zettlemoyer.
\newblock {QLoRA}: Efficient finetuning of quantized {LLMs}.
\newblock \emph{Neural Inf Process Syst}, abs/2305.14314:\penalty0 10088--10115, 23~May 2023.
\newblock \doi{10.48550/arXiv.2305.14314}.

\bibitem[Pandey et~al.(2020)Pandey, Xu, Sholle, Maliakal, Singh, Fatima, Larine, Lee, Wang, van Rosendael, Baskaran, Shaw, Min, and Al'Aref]{pandey2020extraction-k}
Mohit Pandey, Zhuoran Xu, Evan Sholle, Gabriel Maliakal, Gurpreet Singh, Zahra Fatima, Daria Larine, Benjamin~C Lee, Jing Wang, Alexander~R van Rosendael, Lohendran Baskaran, Leslee~J Shaw, James~K Min, and Subhi~J Al'Aref.
\newblock Extraction of radiographic findings from unstructured thoracoabdominal computed tomography reports using convolutional neural network based natural language processing.
\newblock \emph{PLoS One}, 15\penalty0 (7):\penalty0 e0236827, 30~July 2020.
\newblock ISSN 1932-6203.
\newblock \doi{10.1371/journal.pone.0236827}.

\bibitem[Giuffrè and Shung(2023)]{giuffre2023harnessing-r}
Mauro Giuffrè and Dennis~L Shung.
\newblock Harnessing the power of synthetic data in healthcare: innovation, application, and privacy.
\newblock \emph{NPJ Digit. Med.}, 6\penalty0 (1):\penalty0 186, 9~October 2023.
\newblock ISSN 2398-6352.
\newblock \doi{10.1038/s41746-023-00927-3}.

\end{thebibliography}

\newpage
\appendix
\setcounter{subsection}{0}
\setcounter{table}{0}
\setcounter{figure}{0}
\renewcommand{\thesubsection}{S\arabic{subsection}}
\renewcommand\figurename{Supplementary Figure} 
\renewcommand\tablename{Supplementary Table}

\section*{Supplementary materials}
\label{sec:appendix}

\captionsetup[table]{hypcap=false}

\subsection{Descriptions of the four error types}
\label{ssec:error}

- Negation errors occur when negative phrases are incorrectly used or
omitted, leading to a complete reversal of the intended meaning. For
example, the correct statement ``no signs of pneumonia'' in the findings
section might be incorrectly rewritten as ``signs of pneumonia'' in the
impression section.

- Left/right errors involve the incorrect labeling of anatomical sides,
such as mistakenly describing a condition on the left lung when it is
actually on the right.

- Interval change errors refer to any modification within the report
that alters key information or details, such as changes in measurement
values or descriptions of findings that could lead to misinterpretation.

- Transcription errors simulate common mistakes made during the
transcription of radiology reports, including typographical errors,
misspellings, and incorrect medical terminology, which could lead to
misunderstandings.

\subsection{Computing Resources}
\label{ssec:computing}

Experiments were conducted on two NVIDIA A100 GPUs with 80 GB of memory
to ensure proper fine-tuning and inference of the LLMs. Dual A100 GPUs
were used for inferring the Llama-3-70B-Instruct model and its
fine-tuned version. For the fine-tuning and inference of
Llama-3-8B-Instruct and BiomedBERT, a single A100 GPU was used.

\subsection{Descriptions of random and specified prompting
strategies}
\label{ssec:prompting}

The random prompting strategy involved selecting radiology reports
randomly from the training set. In contrast, the specified strategy used
a more deliberate selection, in which reports were empirically chosen
from the training set on the basis of specific instructions. These
strategies allowed evaluation of the impact of different prompting
strategies on model performance.

\subsection{Descriptions of the real-world radiology reports}
\label{ssec:rwd}

These patients with the ICD-9 Code 428 or the ICD-10 Code I50 who were
admitted and discharged between January 2008 and July 2018 were
included. For each report, only the findings and impression sections
were used. The data were reviewed and deidentified in accordance with
IRB protocols.

\subsection{Performance of the baseline models}
\label{ssec:baseline}

After fine-tuning, BiomedBERT achieved moderate performance across the
various metrics. It performed particularly well in detecting left/right
error, with an F1 score of 0.845 (95\% CI, 0.839-0.852), indicating its
strong performance in spatial orientation tasks. However, it showed
worse performance in detecting interval changes and transcription
errors, with F1 scores of 0.426 (95\% CI, 0.418-0.451) and 0.649 (95\%
CI, 0.640-0.654), respectively.

GPT-4-1106-Preview (zero-shot) demonstrated robust performance across
various metrics. It exceled in detecting negation and interval change
errors, with F1 scores of 0.741 (95\% CI, 0.733 - 0.750) and 0.646 (95\%
CI, 0.628 - 0.657), respectively. However, its ability to detect
transcription errors was only moderate, with an F1 score of 0.537 (95\%
CI, 0.524-0.540), indicating room for improvement in this area.

Llama-3-70B-Instruct (vanilla, zero-shot) outperformed the other models.
It achieved an F1 score of 0.676 (95\% CI, 0.666-0.684) for detecting
negation errors and 0.651 (95\% CI, 0.639-0.654) for detecting
left/right errors, showing its limitations in zero-shot settings. The F1
scores for detecting interval change errors and transcription errors
were relatively low, at 0.548 (95\% CI, 0.536-0.562) and 0.277 (95\%
CI, 0.263-0.281), indicating that additional training is needed to
improve its performance in these areas.

\subsection{Performance of each error in real-world scenarios}
\label{ssec:performance}

Based on confirmation by both radiologists, the error detection accuracy
rates for negation, interval changes, and left/right errors were 12/50
(24\%), 15/50 (30\%), and 30/50 (60\%), respectively. Encouragingly, for
transcription errors, the LLMs demonstrated strong error detection
capabilities, achieving a radiologist-confirmed accuracy rate of 42/50
(84\%).

Based on confirmation by at least one radiologist, the error detection
accuracy rates for negation and left/right errors were 31/50 (62\%) and
38/50 (76\%), respectively. Notably, for interval changes and
transcription errors, the radiologist-confirmed detection accuracy rates
were 46/50 (92\%) and 48/50 (96\%), respectively.

\newpage

\captionsetup{justification=raggedright,singlelinecheck=false}

{
\captionof{table}{Examples of radiology reports with errors.}
\label{stab:examples}
\footnotesize
\begin{tabularx}{\textwidth}{l
>{\hsize=.5\hsize\linewidth=\hsize\arraybackslash}X
>{\hsize=1.5\hsize\linewidth=\hsize\arraybackslash}X}
\toprule
Error type & Definition & Example \\
\midrule
Negation & Negation errors occur when negative phrases are
incorrectly used or omitted, leading to a complete reversal of the
intended meaning. & Patient ID: 00033

History: 30 F with sharp left-sided chest pain, evaluate for pleural
effusion

TECHNIQUE: AP chest and abdomen radiograph.

COMPARISON: Radiograph from one day prior.

FINDINGS:

Lines/tubes: None

Mediastinum/Heart: The cardiomediastinal silhouette is
unremarkable.

Lungs/Airways/Pleura: Evidence of pleural effusion. Lung fields are
clear.

Bones/Soft Tissue: No acute osseous abnormalities.

Upper abdomen: Unremarkable

IMPRESSION:

No pleural effusion noted. \\\midrule
Left/Right & Left/right errors involve the incorrect labeling of
anatomical sides, such as describing a condition on the left lung when
it is actually on the right. & Patient ID: 00014

History: 45 M with history of pneumothorax

TECHNIQUE: AP chest and abdomen radiograph.

COMPARISON: Radiograph from one day prior.

FINDINGS:

Lines/tubes: None

Mediastinum/Heart: The cardiomediastinal silhouette is
unremarkable.

Lungs/Airways/Pleura: Small pneumothorax present on the right side.

Bones/Soft Tissue: No acute osseous abnormalities.

Upper abdomen: Unremarkable

IMPRESSION:

Small pneumothorax noted on the left side. \\\midrule
Interval Change & Interval change errors refer to any modification
within the report that alters key information or details. &
Patient ID: 00001

History: 60 F with chest pain and history of hypertension

Technique: AP chest and abdomen radiograph. Standard protocol

Comparison: Radiograph from two days prior.

Findings:

Lines/tubes: None

Mediastinum/Heart: The cardiomediastinal silhouette is
unremarkable.

Lungs/Airways/Pleura: No focal opacities, pneumothorax or pleural
effusion.

Bones/Soft Tissue: No acute osseous abnormalities.

Upper abdomen: Unremarkable

IMPRESSION:

1. Enlargement of the cardiomediastinal silhouette, consistent with
cardiomegaly. \\\midrule
Transcription & Transcription errors simulate common mistakes made
during the transcription of radiology reports. & INDICATION:
\_\_\_F with shortness of breath

// Please evaluate for pneumonia, effusions, edema

TECHNIQUE: PA and lateral views of the chest.

COMPARISON: \_\_\_.

FINDINGS:

The lungs are clear without consolidation, effusion or edema.

Biapical starring, worse on the right is again noted.

The cardiomediastinal silhouette is within normal limits.

No acute osseous abnormalities.

IMPRESSION:

No acute cardiopulmonary process. \\
\bottomrule
\end{tabularx}
}

\newpage

{
\captionof{table}{The ``specified'' examples used for one-shot learning in the experiments.}
\label{stab:specified examples}
\footnotesize
\begin{tabularx}{\textwidth}{l
>{\arraybackslash}X}
\toprule
Error type & Example \\
\midrule
Left/Right & Patient ID: 00002

History: 59 M with recent cough and weight loss, suspected lung tumor

Technique: AP chest and abdomen radiograph.

Comparison: Radiograph from two days prior.

Findings:

Lines/tubes: None

Mediastinum/Heart: The cardiomediastinal silhouette is unremarkable.

Lungs/Airways/Pleura: Mass in the left upper lobe, suggestive of a lung
tumor. No pleural effusion.

Bones/Soft Tissue: No acute osseous abnormalities.

Upper abdomen: Unremarkable

IMPRESSION:

Mass in the right upper lobe, likely a lung tumor. \\
\bottomrule
\end{tabularx}
}

\newpage

{
\captionof{table}{The ``specified'' examples used for four-shot
learning in the experiments.}
\label{stab:specified examples4}
\footnotesize
\begin{tabularx}{\textwidth}{l
>{\arraybackslash}X}
\toprule
Error type & Example \\
\midrule
Negation & Patient ID: 00033

History: 30 F with sharp left-sided chest pain, evaluate for pleural
effusion

TECHNIQUE: AP chest and abdomen radiograph.

COMPARISON: Radiograph from one day prior.

FINDINGS:

Lines/tubes: None

Mediastinum/Heart: The cardiomediastinal silhouette is unremarkable.

Lungs/Airways/Pleura: Evidence of pleural effusion. Lung fields are
clear.

Bones/Soft Tissue: No acute osseous abnormalities.

Upper abdomen: Unremarkable

IMPRESSION:

No pleural effusion noted. \\\midrule
Left/Right & Patient ID: 00014

History: 45 M with history of pneumothorax

TECHNIQUE: AP chest and abdomen radiograph.

COMPARISON: Radiograph from one day prior.

FINDINGS:

Lines/tubes: None

Mediastinum/Heart: The cardiomediastinal silhouette is unremarkable.

Lungs/Airways/Pleura: Small pneumothorax present on the right side.

Bones/Soft Tissue: No acute osseous abnormalities.

Upper abdomen: Unremarkable

IMPRESSION:

Small pneumothorax noted on the left side. \\\midrule
Interval Change & Patient ID: 00001

History: 60 F with chest pain and history of hypertension

Technique: AP chest and abdomen radiograph. Standard protocol

Comparison: Radiograph from two days prior.

Findings:

Lines/tubes: None

Mediastinum/Heart: The cardiomediastinal silhouette is unremarkable.

Lungs/Airways/Pleura: No focal opacities, pneumothorax or pleural
effusion.

Bones/Soft Tissue: No acute osseous abnormalities.

Upper abdomen: Unremarkable

IMPRESSION:

1. Enlargement of the cardiomediastinal silhouette, consistent with
cardiomegaly. \\\midrule
Transcription & INDICATION: \_\_\_F with shortness of breath

// Please evaluate for pneumonia, effusions, edema

TECHNIQUE: PA and lateral views of the chest.

COMPARISON: \_\_\_.

FINDINGS:

The lungs are clear without consolidation, effusion or edema.

Biapical starring, worse on the right is again noted.

The cardiomediastinal silhouette is within normal limits.

No acute osseous abnormalities.

IMPRESSION:

No acute cardiopulmonary process. \\
\bottomrule
\end{tabularx}
}

\newpage

{
\captionof{table}{Typical examples identified by fine-tuned
Llama-3-70B-Instruct and GPT-4-1106-Preview and verified by two
physicians.}
\label{stab:typical examples}
\footnotesize
\begin{tabularx}{\textwidth}{l
>{\arraybackslash}X}
\toprule
Error type & Example \\
\midrule
Negation & Clinical statement: Syncope.

Technique: Single AP chest dated 04/03/2010 at 07:32.

No prior films available for comparison.

Findings:

The overall heart size is within normal limits. The lungs are well
expanded and clear.

Impression:

Cardiomegaly with atherosclerotic changes thoracic aorta. \\\midrule
Left/Right & Technique: PA and lateral views of the chest.
Supine and upright views of the abdomen.

Comparison: 7/24/2009 (chest). No prior abdominal radiographs for
comparison.

Findings:

Chest: Minimal patchy opacity at the right lung base may reflect
atelectasis or consolidation.

There is no pleural effusion or pneumothorax. The cardiomediastinal
silhouette remains enlarged.

Abdomen: There is no pneumoperitoneum. No dilated bowel is seen. Air
is seen in the colon.

Impression:

1. Enlarged cardiomediastinal silhouette.

2. Minimal left basilar atelectasis or consolidation.

3. Nonobstructive bowel gas pattern. \\\midrule
Interval Change & Clinical Statement: Abnormal chest sounds.

Technique: AP view of the chest.

Comparison: January 21, 2010.

Findings:

Persistent near-complete opacification on the left hemithorax with
associated leftward cardiomediastinal shift, compatible with volume
loss, is unchanged since the prior study. The right lung remains clear.
Metallic rods and screws transfix the thoracic spine.

Impression:

Since January 21, 2010:

1. Near-complete opacification of the left hemithorax with
associated leftward cardiomediastinal shift compatible with volume loss,
increased since the prior study. Evaluation for superimposed focal
consolidation is limited. \\\midrule
Transcription & Technique: Portable AP chest radiograph.

Comparison: June 1, 2010.

Findings: The lungs are clear. No pleural effusion or pneumothorax
is identified.

The cardiomediastinal silhouette is within normal limits.

Radiopaque densities project across the chest.

Impression:

1. No acute cardiopulmonary disease.

2. Densities projecting over the chest may be related to
patient's clothing, clinically correlate with repeat
examination as clinically indicated. \\
\bottomrule
\end{tabularx}
}

\end{document}